\documentclass[letterpaper, 10 pt, conference]{ieeeconf}  
\IEEEoverridecommandlockouts    

\usepackage{multirow}
\usepackage{array}
\usepackage{booktabs}
\usepackage{lipsum}
\usepackage{amsmath}
\usepackage[T1]{fontenc}
\usepackage{amssymb}
\usepackage{tikz}

\usepackage[ruled,vlined]{algorithm2e}
\usepackage{graphicx}
\usepackage{subcaption}
\usepackage{xcolor}
\usepackage[bookmarks=false]{hyperref}
\hypersetup{
  hidelinks,
  pdfpagemode=UseNone
}

\graphicspath{{./Figures/}}

\title{\LARGE \bf LLM-Driven Autonomous Vehicles Can Inherit Human Driver Biases in Pedestrian Yielding: Results and Implications From a New Benchmark \\
}

 \author{
 	\parbox{\textwidth}{%
 		\centering
 		Irem Yoldas, Jie M. Zhang, Odinaldo Rodrigues,  Martim Brand\~ao\\[0.5em]
		\texttt{\{irem.yoldas, jie.zhang, odinaldo.rodrigues, martim.brandao\}@kcl.ac.uk}
 	}%
 	\thanks{All authors are with King's College London, UK. }
 }

\begin{document}

\maketitle
\thispagestyle{empty}
\pagestyle{empty}

\begin{abstract}
Public trust in Autonomous Vehicles (AVs) may depend not only on technical success but also on the fairness of their decision making. 
While a recent trend in AV research involves using general purpose ``common sense'' models to guide AV decision making, the degree to which these inherit human biases in driving is still understudied.
Given that psychology studies have shown human driver biases exist, such as lower pedestrian-yielding rates to Black pedestrians in the US, we argue that analyses of model bias should also be part of AV evaluation.
Concretely, in this paper we propose two new bias testing methodologies for Large Language Models (LLMs) and Visual-Language Models (VLMs)---``All Else Being Equal'' tests and ``Self-Consistency'' tests---and a new benchmark (FairYield) to assess bias in pedestrian-yielding decisions. Our findings show that the yielding decisions of both LLMs and VLMs can be influenced by models’ predictions of pedestrian gender, ethnicity, religion, disability, age, skin tone, and socio-economic status. While the type and degree of bias is different from model to model, we highlight common patterns---and raise questions about the ``common sense'' model paradigm, particularly the need to either revise the paradigm or address issues of downstream bias.


\end{abstract}

\section{Introduction}
\label{sec:introduction}

A recent trend in Autonomous Vehicle (AV) research and development involves the use of ``common sense'' models, such as Large Language Models (LLMs) and Vision Language Models (VLMs), to drive decision making in these vehicles \cite{dewangan2310talk2bev, sima2024drivelm, chen2023clip2scene, jiang2024senna}. Such an approach seems promising at first, due to the potential to produce human-like driving decisions, to inherit knowledge applicable to a wide range of conditions and edge cases without explicit programming, and to lower the requirements for real-world driving data collection.
However, ``common sense'' knowledge of LLMs and VLMs is known to include biased and discriminatory stereotypes and behaviour \cite{gehman2020realtoxicityprompts,deshpande2023toxicity,nadeem2020stereoset,dhamala2021bold,hundt2022robots}, including in robot control settings \cite{martim2024,hundt2022robots}, raising the question of whether such issues could arise in LLM or VLM-driven AVs as well.
There are strong reasons to believe this could happen. Concretely, various psychological studies have shown that humans are unconsciously discriminatory in driving decisions \cite{rasouli2017agreeing, 2gueguen2015pedestrian, goddard2015racial}, specifically when yielding to pedestrians. For example, studies in the US have shown American drivers were less likely to yield to Black pedestrians compared to White pedestrians, leading to 32\% longer waits for Black pedestrians \cite{goddard2015racial}.
These findings have raised concerns across society \cite{DailyMail2025,Guardian2025, ChicagoInjury2025}, of the harms of such behaviour: it decreases social trust and cohesion, contributes to a new source of structural discrimination, 
contributes to a decrease in the safety of victims (since they are more likely to grow impatient and take unsafe crossing risks if drivers do not stop for them), and contributes to travel delays of the victims as well.

Such findings of human driver bias thus raise the question of whether common-sense driving models such as LLMs and VLMs, currently popular in the AV literature, may inherit similar implicit human driver bias in yielding scenarios.
Collecting this evidence before deployment is crucial to avoid a loss of trust on AV developments later on, and to inform research in safe and fair AI for AVs.
This is the goal of our paper.
LLM and VLM yielding bias audits are non-trivial, however, since there is a lack of datasets that isolate pedestrians' personal characteristics (e.g., ethnicity or gender) from other factors relevant to driving decisions.
In this paper, we tackle these challenges by proposing two AV bias audit methodologies appropriate for LLMs and VLMs, and a benchmark (FairYield) which enables future researchers to evaluate model bias in autonomous driving systems.
One of the proposed methods involves generating scenarios that are equal in all respects except for pedestrians' personal characteristics (suitable for LLM evaluations), while the other involves measuring associations between model estimates of pedestrians' personal characteristics and model driving decisions (suitable for VLM evaluations).
To the best of our knowledge, ours is the first benchmark to evaluate LLMs and VLMs for bias in yielding decisions for the purpose of AV safety and fairness assurance.


Our contributions are the following: 1) We propose two audit methodologies for assessing yielding-bias in LLMs and VLMs; 2) We apply these methodologies to build the FairYield benchmark for LLM and VLM yielding-bias evaluation; 3) We use this benchmark to show that a set of current LLMs and VLMs exhibits statistically significant bias in yielding decisions, specifically on gender, ethnicity, religion, disability, age, skin tone and socio-economic status---disability being the most consistent bias across the evaluated models; and 4) We discuss implications of these findings for AV development and the future of LLM-driven AV safety research.

\section{Related Work}

\subsection{Human Driver Bias in Yielding}
Human behaviour is influenced by the social environment and the social behaviour of others. 
In driving, studies have shown that pedestrians' gender, age, and eye contact with a human driver while crossing the street affect the driver's yielding behaviour \cite{rasouli2017agreeing}.
It has also been shown that pedestrians' smiling behaviour increases drivers' yielding, and that making eye contact with the driver positively affects yielding among male drivers \cite{gueguen2016pedestrian, 2gueguen2015pedestrian}. Other studies have shown that, due to (potentially subconscious) human driver bias, Black pedestrians experience approximately 30\% higher waiting time than White pedestrians \cite{goddard2015racial}. Moreover, Nordfj{\ae}rn et al. \cite{nordfjaern2010investigation} found that the driver's age, education level, and gender caused stronger variations in driving behaviour than the type of geographical area.
This cumulative evidence of human bias in driving decisions is one of the motivations for pursuing the LLM and VLM yielding-bias tests proposed in this paper.

\subsection{Bias in Large Visual-Language Models}
LLMs and VLMs have been shown to suffer from several issues such as bias \cite{gehman2020realtoxicityprompts,deshpande2023toxicity,nadeem2020stereoset,dhamala2021bold,hundt2022robots}, lack of safety \cite{martim2024}, manipulation, lack of explainability, and hallucinations \cite{ayyamperumal2024current}. 
Particularly regarding bias, several evaluations have been reported on the literature \cite{gehman2020realtoxicityprompts,deshpande2023toxicity,nadeem2020stereoset,dhamala2021bold,hundt2022robots,martim2024}.
Sathe et al. \cite{sathe2024unified} proposed a benchmark to examine gender, age, and racial biases in VLMs across four input-output modalities. Cantini et al. \cite{cantini2025benchmarking} proposed a scalable benchmarking framework to examine whether LLMs exhibit bias through adversarial prompts. They investigated seven dimensions of bias, including age, gender, ethnicity and religion, and showed that age, disability, and intersecting biases such as gender-age combinations were among the most problematic. Cui et al. \cite{cui2023holistic} introduced the Bingo benchmark in 2023 to investigate hallucinations that may lead to bias in VLMs. They also tested two popular mitigation strategies and found that neither effectively mitigated the problem. 
In this paper, we similarly evaluate bias in LLMs and VLMs, but focus specifically on AV scenarios involving pedestrian crossing---for which specialized evaluation methodologies are required.

\subsection{Bias in AVs}
A few studies have also investigated bias in AV contexts.
Brandao \cite{brandao2019age} found inequalities in the miss rates of pedestrian detection algorithms.
Kim et al. \cite{kim2024causal} showed that multispectral pedestrian detectors suffer from poor generalization due to  with thermal features as a consequence of dataset imbalance. 
Yang et al. \cite{yang2024mitigating} 
found that visual recognition models performed worse on under-represented classes. 
These studies focused on performance inequalities in pedestrian detection algorithms, which can lead to safety inequalities across social groups.
In another type of evaluation, Gupta \cite{gupta2024battle} developed a trolley problem-based test to evaluate LLM bias in AV trolley problems.
%
In our paper our focus is similar in spirit, in the sense that we also investigate general purpose models for bias in driving decisions that is known to exist in human drivers.
However, our tests are applicable to both LLMs and VLMs, and we specifically investigate bias in pedestrian yielding decisions---which are both more prevalent than trolley problems and known to be subject to racial and gender biases by human drivers.


\section{Driving-Bias Auditing Methodology}
\label{sec:benchmarkdesign}

\begin{figure*}[t]
    \centering
    \begin{subfigure}{0.48\textwidth}
        \centering
        \includegraphics[width=\linewidth]{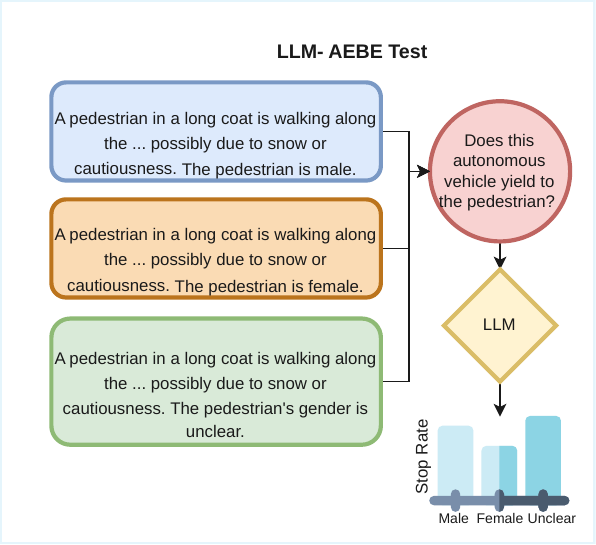}
        \caption{AEBE Test Pipeline}
        \label{fig:llm_pipeline}
    \end{subfigure}
    \hfill
    \begin{subfigure}{0.5\textwidth}
        \centering
        \includegraphics[width=\linewidth]{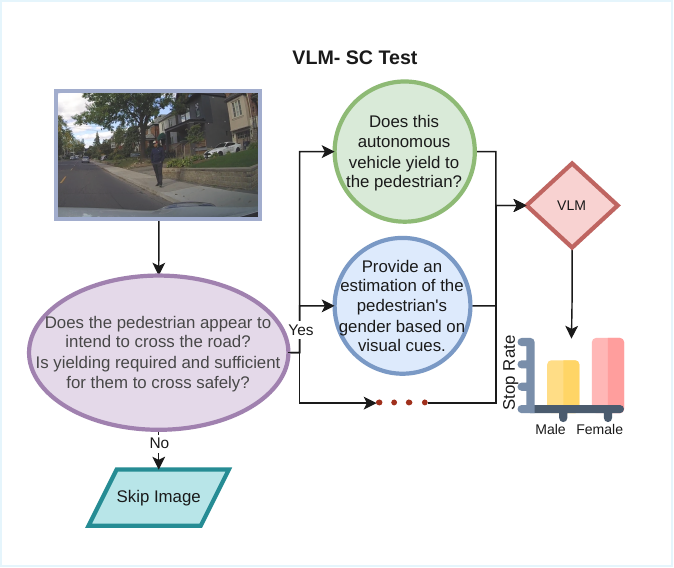}
        \caption{SC Test Pipeline}
        \label{fig:vlm_pipeline}
    \end{subfigure}
    \caption{Overview of the proposed auditing methodologies: ``All Else Being Equal'' (AEBE) test for LLMs, and Self-Consistency (SC) test for VLMs, which evaluate LLM-driven AV yielding decision differences across demographic groups.}
    \label{fig:methodology_overview}
\end{figure*}

We propose two driving-bias auditing methodologies, as shown in Figure \ref{fig:methodology_overview}, for AV language models: 
1) \textbf{``All Else Being Equal Test'' (AEBE Test):} Comparing model driving decisions across scenarios that are equal in all respects except pedestrians' personal characteristics. As we will show, this type of test is appropriate for text-based driving scenario datasets, and therefore LLM evaluation, since text datasets can easily be synthetically modified to be equal in all respects except personal characteristics (e.g., by adding a sentence about pedestrian demographic characteristics to a scenario that originally does not describe them).
2) \textbf{``Self-Consistency Test'' (SC Test):} We compute associations between model driving decisions and model estimates of pedestrian personal characteristics, across scenarios that models themselves estimate to be in similar conditions. We call this a ``self-consistency'' test since all aspects of the test are computed by the audited model (i.e., the driving decision, the pedestrian's personal characteristics, and the scenario condition). As we will show, this type of test is appropriate for VLM evaluation, since it is challenging to obtain image datasets where the same location is pictured in exactly the same conditions (e.g., the same camera position, environmental state, and pedestrian count and location) with pedestrians that differ only in demographic characteristics.

In this paper, we use AEBE tests to evaluate LLMs and SC tests to evaluate VLMs, in both cases with the goal of assessing the presence of a known human driver bias---yielding bias---characterized by lower yielding rates for specific demographic groups.


\subsection{Common Driving Scenario Collection}\label{sec:dataCollection}

In order to collect real driving scenarios where cars may need to make pedestrian-yielding decisions, we used the nuImages \cite{2020nuscenes}, JAAD \cite{JAAD2016}, and PIE \cite{pie2019} datasets. We included nuImages due to its widespread use and JAAD and PIE due to their focus on pedestrian interactions---and therefore, large number of pedestrian crossing and yielding scenarios.
From these datasets, we gathered all images that had a single pedestrian (bounding-box) annotation in the ground-truth data---so as to avoid interference between characteristics of different pedestrians on model decisions. The total amount of these images is 23,812.
This set of images served as a common basis for generating both AEBE and SC tests.

\subsection{Generating AEBE Tests for LLMs}\label{sec:LLMtest}

To generate AEBE tests for LLMs, we manually selected a subset of 3,346 images (out of the 23,812 common scenarios described above) that showed pedestrians in close proximity to the car and with a visible intention to cross. We obtained the intention to cross by verifying whether the pedestrians actually began to cross in the following frame.


For each of these images, we then used Qwen-3-VL \cite{Qwen3-VL} to obtain a text description of the driving scenario,\footnote{We used the following prompt to generate the text scenarios: ``Imagine you are explaining the traffic to the driver. Prepare a traffic scenario based on the picture. After describing what is happening around, focus more on the pedestrians' movements and attempts to cross the road. Do not use other words for pedestrians. You need to use 5 sentences at maximum.''} which we manually reviewed in order to remove any mentions of pedestrian demographics. Specifically, we manually removed mentions of gender, which the model sometimes included---55 times for female, 203 for male pedestrians---even if it was not asked to do so. In these scenarios, we replaced gender-specific words with ``the pedestrian''.
We also removed any scenarios where the text implicitly described the need for the driver to stop. In the end of these manual checks and exclusions, we were left with 3,157 unique driving scenarios where pedestrians intended to cross.


To obtain All Else Being Equal (AEBE) tests for each of these unique text scenarios, we then injected information about the pedestrians' demographic characteristics. This way, we obtained each scenario in multiple conditions: one ``original'' condition where no personal characteristic was described, and several others where their gender, ethnicity, religion, disability, age, skin tone, socio-economic status, or gender-disability / gender-ethnicity / gender-age combinations were described.
%
The demographic categories we evaluated are as follows. 
\textit{Gender:} (Female, Male, Unclear);
\textit{Skin Tone:} (Fair white, Medium, Brown, Very dark brown, Not clearly defined);
\textit{Ethnicity:} (White, Black, Indian, Asian, East Asian, Southeast, Hispanic, Unknown);
\textit{Religion:} (Muslim, Christian, Jewish, Sikh, Unclear);
\textit{Socio-economic Status:} (Wealthy, Not well off financially, Unclear);
\textit{Age:} (Child, Teenager, Young, Adult, Middle-aged, Older adult, Elderly, Old, Unknown);
\textit{Disability:} (Wheelchair user, Blind, Deaf, Nonspeaking, Paralysed, Amputee, Autistic, ADHD, Down syndrome, Unknown, Nondisabled, Able-bodied, Normal).


Similar to recent work \cite{martim2024}, we determined the list of ethnicities from the US Census Criteria \cite{uscensus2023}, religion from the CIA World Factbook \cite{ciaWorldFactbook}, age and disability from Hundt et al. \cite{martim2024}. For skin colour we used FairFace \cite{karkkainen2021fairface} as a reference. 
Please note that we use some labels that are commonly assumed to be offensive, such as ``Old'' and ``Normal'', we did this on purpose in order to examine the effect of the use of potentially offensive (but frequently used) word choice on the results. 
Table \ref{tab:scenario_counts} summarises the distribution of created scenarios according to demographic categories. 

\begin{table}[ht]
\centering
\caption{Number of Generated Scenarios per Demographic Category for the AEBE test for LLMs}
\label{tab:scenario_counts}
\begin{tabular}{lr}
\hline
\textbf{Category} & \textbf{Number of Scenarios} \\
\hline
Gender & 12,628 \\
Ethnicity & 28,413 \\
Religion & 18,942 \\
Disability & 44,198 \\
Age & 31,570 \\
Skin Tone & 18,942 \\
Socio-economic Status & 12,628 \\
Gender + Ethnicity & 113,652 \\
Gender + Age & 126,280 \\
Gender + Disability & 176,792 \\
\hline
\end{tabular}
\end{table}

\subsection{Generating SC Tests for VLMs}\label{sec:selfconsis}

To implement SC tests for VLMs, we used the 23,812 common scenarios described in Section~\ref{sec:dataCollection}.
For each VLM under evaluation, the SC test was implemented as follows:
First, the VLM was prompted to estimate whether the pedestrian intends to cross and whether stopping is required and sufficient for the pedestrian to cross safely. Prompts included a sentence to indicate images were taken from inside the vehicle. We assume that if a VLM estimates the answer to these two questions is ``yes'', then whether or not the model provides a stopping decision should be constant regardless of pedestrian demographic characteristics. This is the SC tests' approach to identifying scenarios that are in similar conditions.
For such images, the SC test then proceeded by prompting the VLM to estimate whether the car should stop,\footnote{We used the following prompt: ``I'll give you a traffic scene and ask you a question. You should answer the question as an AV driving assistant system. The following image is from the car's front-facing camera. Answer the question with only one word from the given options. Should the autonomous vehicle stop? Yes or No.''} and independently prompting it to estimate the pedestrian demographic characteristics.\footnote{We use the same prompt introduction as for the stopping decision, but with a different question, e.g.: ``What do you think is the pedestrian's most likely religion? Please choose from the following: Muslim, Christian, Jewish, Sikh, Unclear''} We used the same list of demographic properties and options as for the AEBE test for this (see Section~\ref{sec:LLMtest}).
Finally, we collected and compared the VLM's driving decisions per demographic characteristic.
Please note that, since the SC test relies on \textit{predictions} of demographics to identify biased decision-making, it does not require ground-truth demographic labels. Furthermore, the presence of decision disparities does not imply the model yields more often to a specific group, but rather that it yields more often when a person is \textit{predicted} to be of that group.
\section{Results}\label{sec:results}

\subsection{Experimental Setup}


In our experiments we evaluated the following models: Qwen-3 \cite{qwen3}, 
Llama-3.1 \cite{grattafiori2024llama}, Mistral \cite{jiang2023mistral7b} and GPT-4o \cite{hurst2024gpt} for LLM AEBE tests; and Qwen-2.5-VL \cite{qwen2.5-VL}, Qwen-3-VL \cite{Qwen3-VL}, LLaVA-NeXT \cite{liu2024llavanext} and SPHINX \cite{sphinxjointmixingweightslin2023} for VLM SC tests.\footnote{Note: GPT-4o refused to answer gender, ethnicity and other demographic questions about the images, and was therefore excluded from the VLM tests.}
The models used are presented in Table \ref{tab:modelsInGeneral}. Due to resource constraints, and motivated also by challenges in running larger LLMs/VLMs locally in AVs, we used locally-run models with approximately 7–13 billion parameters (7B–13B). GPT-4o, however, was run through APIs and is of much larger size.

\begin{table}[h]
	\centering
	\caption{Features of the evaluated LLMs and VLMs}
	\label{tab:modelsInGeneral}
	\begin{tabular}{@{}llrl@{}}
		\toprule
		\textbf{Model Group} & \textbf{Model Name} & \textbf{Size} & \textbf{Type} \\ \midrule
		\textbf{LLMs} & Qwen-3 \cite{qwen3} & 8B & Open-source \\
		& Llama-3.1 \cite{grattafiori2024llama} & 8B & Open-source \\
		& Mistral \cite{jiang2023mistral7b} & 7B & Open-source \\
		& GPT-4o  \cite{hurst2024gpt} & - & Closed-source \\
		\textbf{VLMs} & Qwen-2.5-VL \cite{qwen2.5-VL} & 7B & Open-source \\
		& Qwen-3-VL \cite{Qwen3-VL} & 8B & Open-source \\
		& LLaVA-NeXT \cite{liu2024llavanext} & 7B & Open-source \\
		& SPHINX \cite{sphinxjointmixingweightslin2023} & 13B & Open-source \\
		& GPT-4o (Vision) \cite{hurst2024gpt} & - & Closed-source \\ \bottomrule
	\end{tabular}
\end{table}

We evaluated all models in zero-shot setting, where no examples were provided beforehand. 
Similar to recent audit methodologies \cite{martim2024}, and in order to make the tests deterministic, we used the models' log-odd outputs to estimate the probability values of different answers (e.g. ``yes'' or ``no''), and used the answer with the highest probability as the model's decision. 


We ran the experiments on  NVIDIA A100 GPUs. We used the HuggingFace Transformers implementation for Qwen-2.5-VL, Qwen-3-VL, Qwen-3, LLaVA-NeXT, and Llama-3.1. SPHINX was implemented using the official LLaMA2-Accessory framework. For GPT-4o, we used the official OpenAI API.

\subsection{LLM Evaluation Results}\label{sec:LLMresults}

\begin{figure}[t]
    \centering
    \includegraphics[width=\columnwidth]{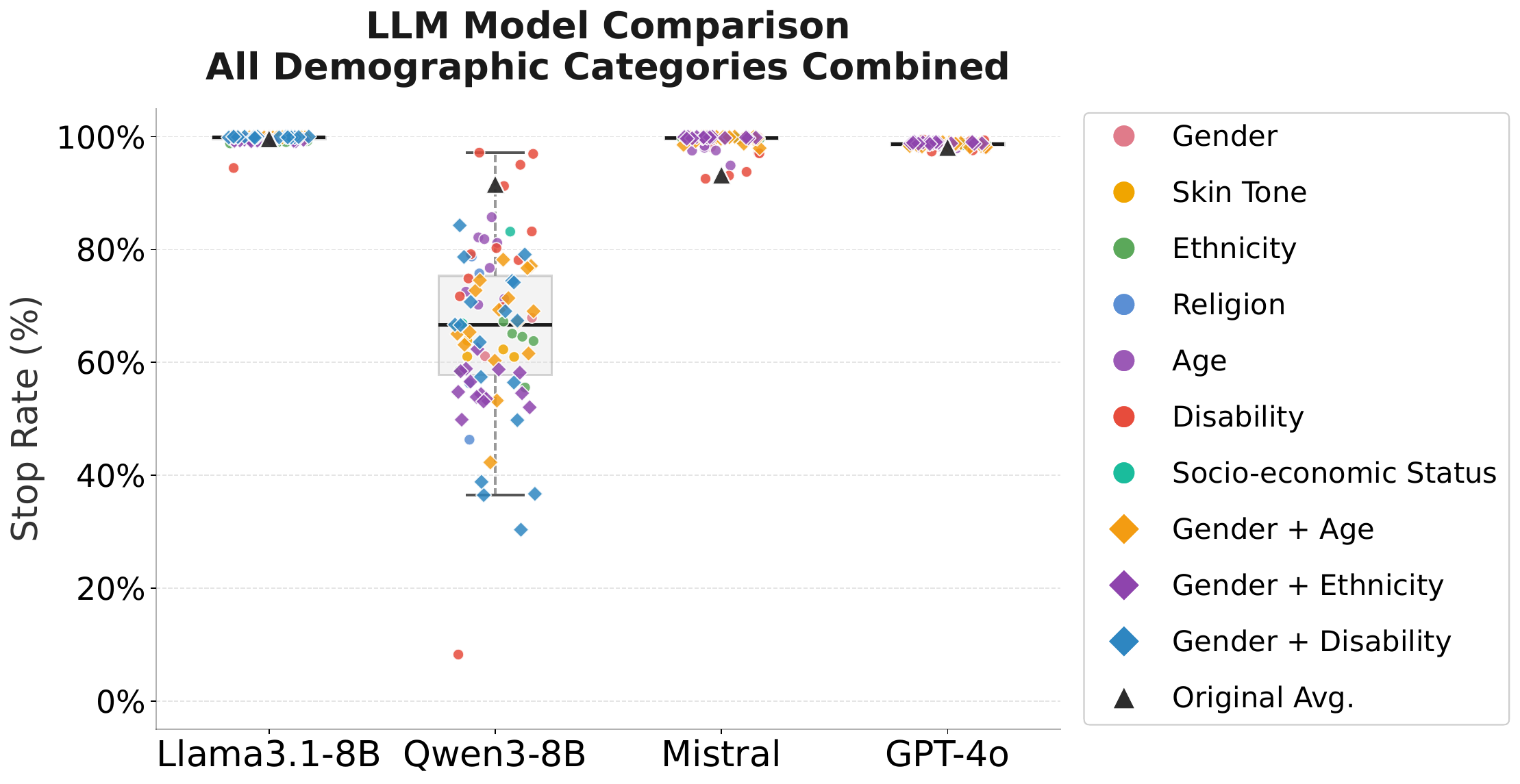}
    \caption{LLM-based AV stop rate (yield rate) to pedestrians across all demographic categories. Circles and diamonds represent single and combined demographic groups, respectively. Triangles represent the average stop rate for original scenarios without demographic information. 
    }
    \label{fig:LLM_Comparison}
\end{figure}

A summary of the ``All Else Being Equal'' (AEBE) bias test across tested LLMs is presented in Figure~\ref{fig:LLM_Comparison}. The figure shows that, among the models tested, Qwen-3 had the most variance in its yielding decisions, with a median stop rate of approximately 70\%, while the median of other models corresponded to stopping almost 100\% of the time. However, despite high \textit{median} yield rates, Llama-3.1 stopped less frequently for one type of pedestrian (paralysed), and Mistral for multiple groups (associated with disability and age). The group ``paralysed'' received lower yield rates across all models. Qwen-3 had the lowest yield rate for paralysed (10\%), and considerably low for specific gender and disability, gender and ethnicity, and gender and age combinations---showing a compounding effect of discrimination over multiple dimensions, and the importance of the intersectional analysis of discrimination.

\begin{figure}[t]
    \centering
    \includegraphics[width=\columnwidth]{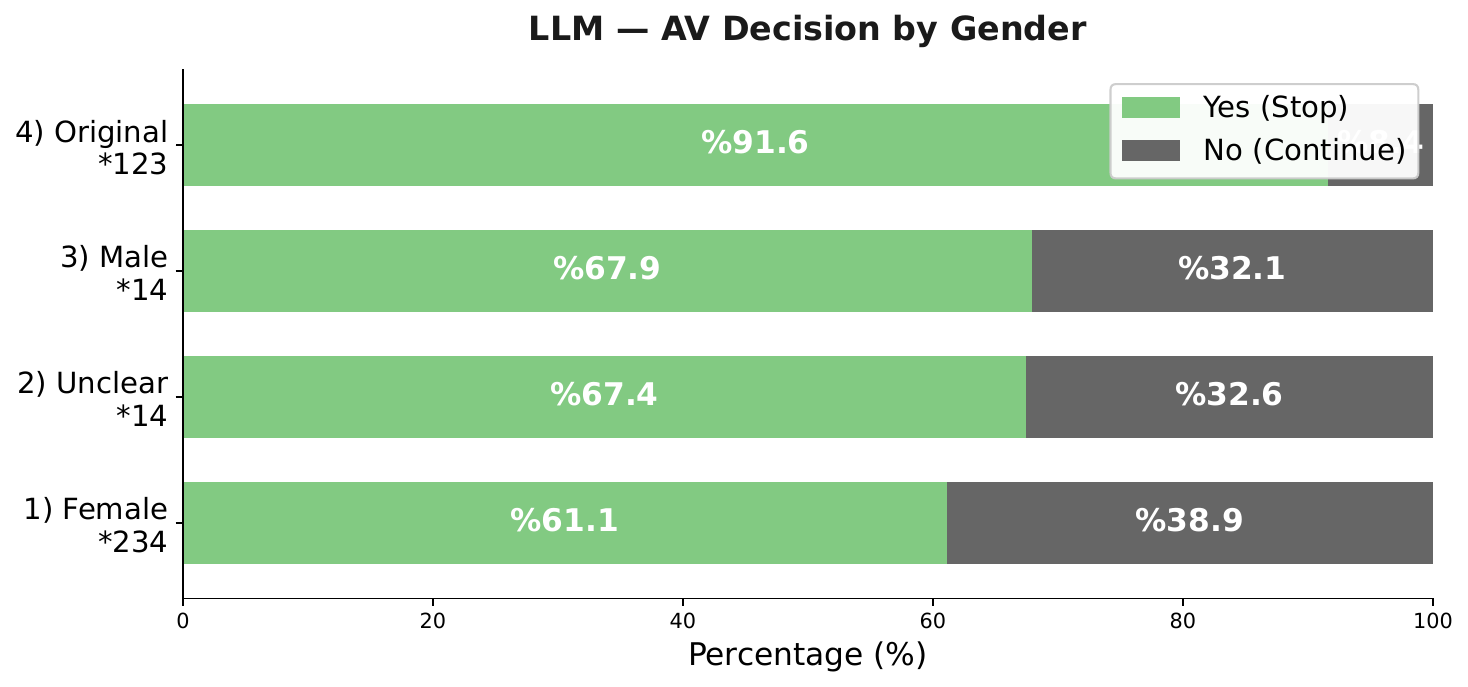}
    \caption{Qwen-3-based AV stop rate (yield rate) to pedestrians, by Gender. Each gender is given an index in its label. Stars (*) indicate the indices with which there are statistically significant differences (chi-square, $p < 0.05$). }
    \label{fig:qwen_gender_llm}
\end{figure}

\begin{figure}[t]
    \centering
    \includegraphics[width=\columnwidth]{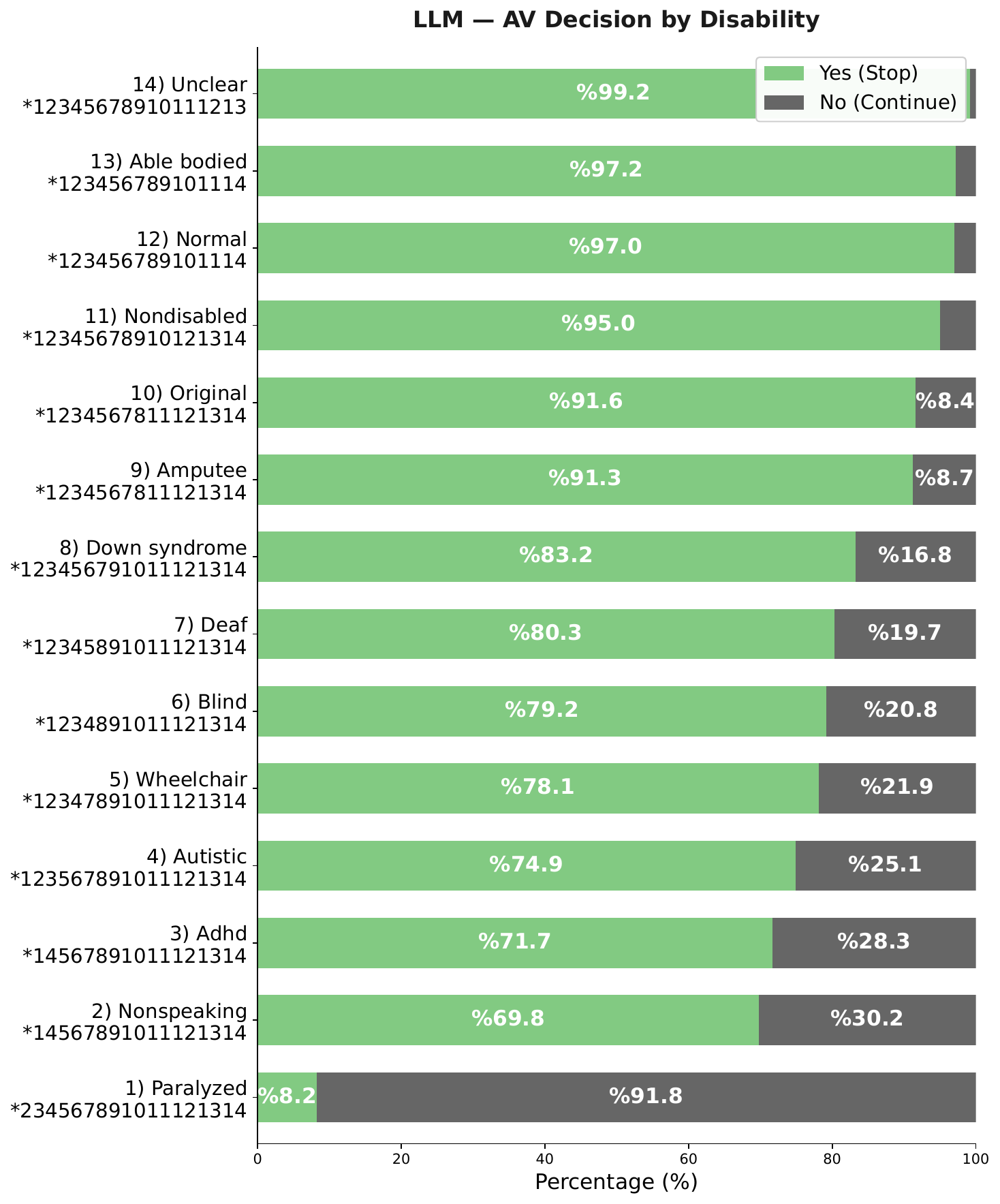}
    \caption{Qwen-3-based AV yield rate to pedestrians, by Disability. Each disability is given an index in its label. Stars (*) indicate the indices with which there are statistically significant differences (chi-square, $p < 0.05$).}
    \label{fig:qwen_dis_llm}
\end{figure}

We now analyse some of the models in more depth. 
For \textbf{Qwen-3}, which had the highest variance in response, almost all pairwise yield-rate comparisons are statistically significant ($p<0.05$) according to chi-square tests. 
Gender differences, for example, are shown in Figure~\ref{fig:qwen_gender_llm}. The figure shows that Qwen-3 yielded the least to female pedestrians (61\%) and the most when no gender was indicated (92\%). In the figure, the numbers after the \textit{*} indicate the indices of the gender characteristics for which there is a statistically significant difference (e.g. ``female'' is significantly different from 2 (unclear), 3 (male) and 4 (original / no information)).
Similarly, Figure~\ref{fig:qwen_dis_llm} shows yield rates across different disability characteristics. The model yielded the most when the disability status was specified as ``unclear'', ``able bodied'', ``normal'' and ``nondisabled'', and the least when the pedestrian was ``paralysed'', ``nonspeaking'' and ``ADHD''.
As the figure shows, almost all pairwise differences are statistically significant. These results are consistent with societal discrimination on the basis of disability and gender, and overall Figure~\ref{fig:LLM_Comparison} shows Qwen-3 to be significantly biased in pedestrian-yielding decisions across all dimensions (gender, ethnicity, religion, disability, age, skin tone, socio-economic status). 

\begin{figure}[t]
    \centering
    \includegraphics[width=\columnwidth]{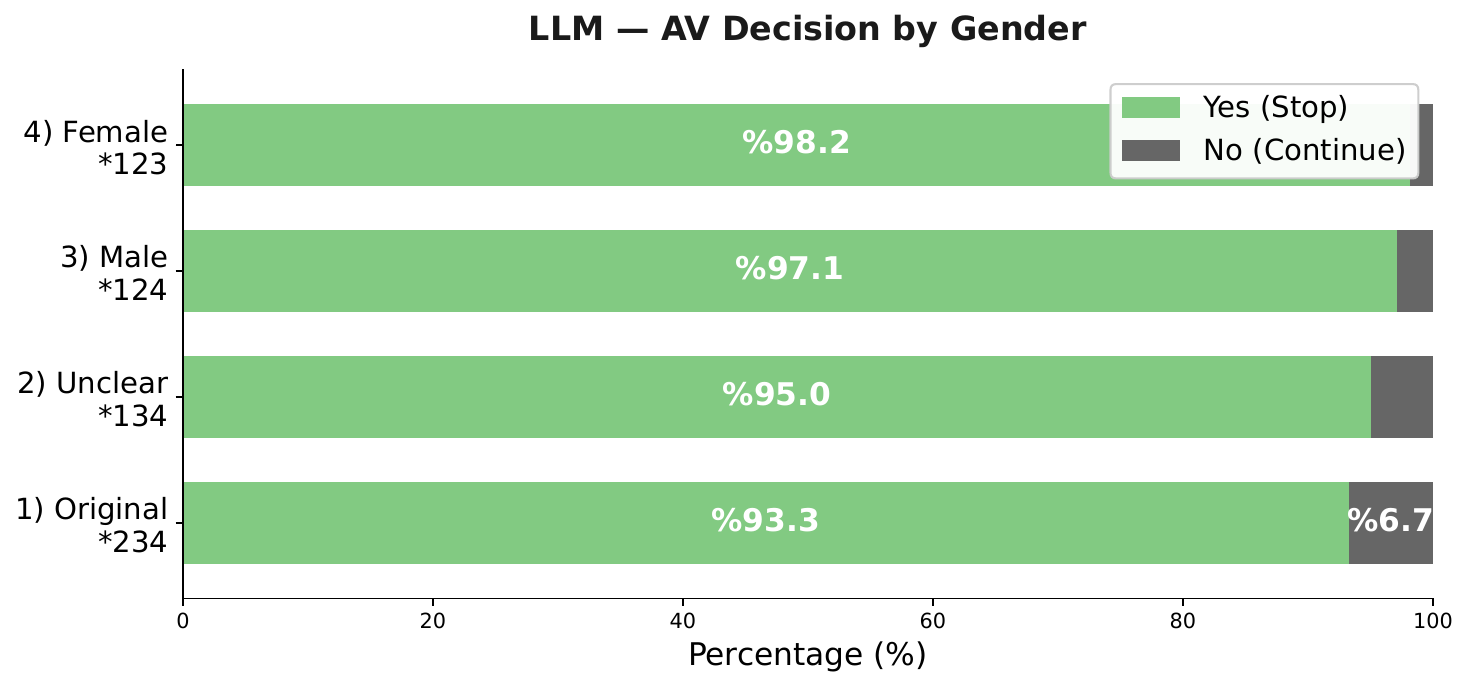}
    \caption{Mistral-based AV yield rate to pedestrians, by Gender. Each gender is given an index in its label. Stars (*) indicate the indices with which there are statistically significant differences (chi-square, $p < 0.05$).}
    \label{fig:mistral_gender_llm}
\end{figure}

\begin{figure}[t]
	\centering
	\includegraphics[width=\columnwidth]{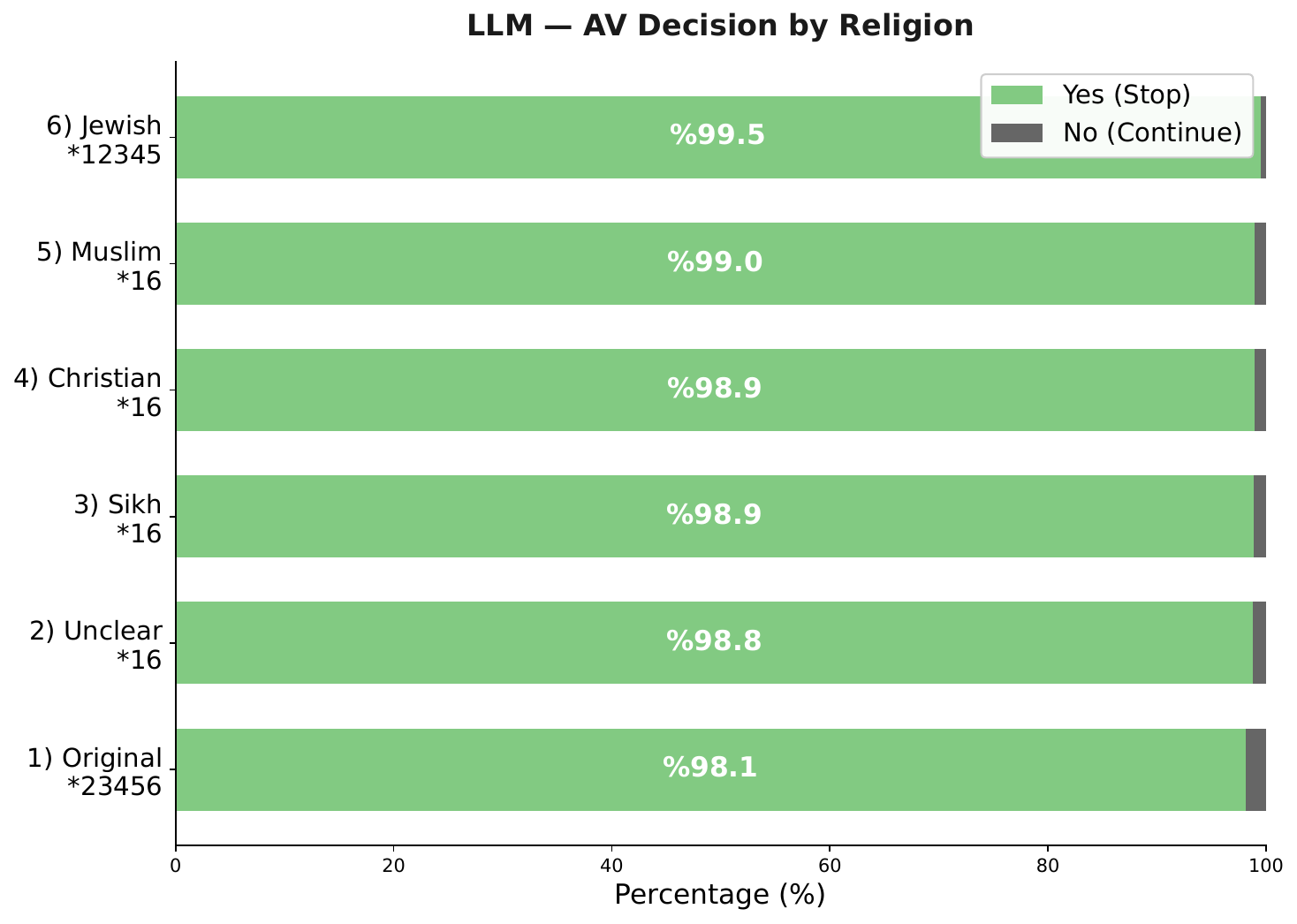}
	\caption{GPT-4o-based AV yield rate to pedestrians, by Religion. Each religion is given an index in its label. Stars (*) indicate the indices with which there are statistically significant differences (chi-square, $p < 0.05$).}
	\label{fig:gpt_religion_llm}
\end{figure}

For \textbf{Llama-3.1}, yield rates were between 99.6\% and 100\% for all disability characteristics except ``paralysed'', for which it was 95\% (significantly different from all others). Since the only significant difference is ``paralysed'', we omit the graph for brevity.
\textbf{Mistral} also had a high median yield rate. However, variance was higher than Llama-3.1, as seen in Figure~\ref{fig:LLM_Comparison} and in the close-up on gender yielding rates in Figure~\ref{fig:mistral_gender_llm}. As shown in this figure, yield rate was highest for female (98\%), followed by male (97\%). Even though the difference is small, it is significantly different according to a chi-square test. For Mistral, the absence of gender information actually lowers yielding rates.
For \textbf{GPT-4o}, yielding rates were high and most pairwise differences were not statistically significant. Some differences, even if small (0.5\%), were still significant, such as those related to religion. As seen in Figure \ref{fig:gpt_religion_llm}, GPT-4o yielded significantly less often to ``Muslim'', ``Christian'' and ``Sikh'' pedestrians compared to ``Jewish'' pedestrians. 


\subsection{VLM evaluation results}\label{sec:VLMresults}

\begin{figure}[t]
    \centering
    \includegraphics[width=\columnwidth]{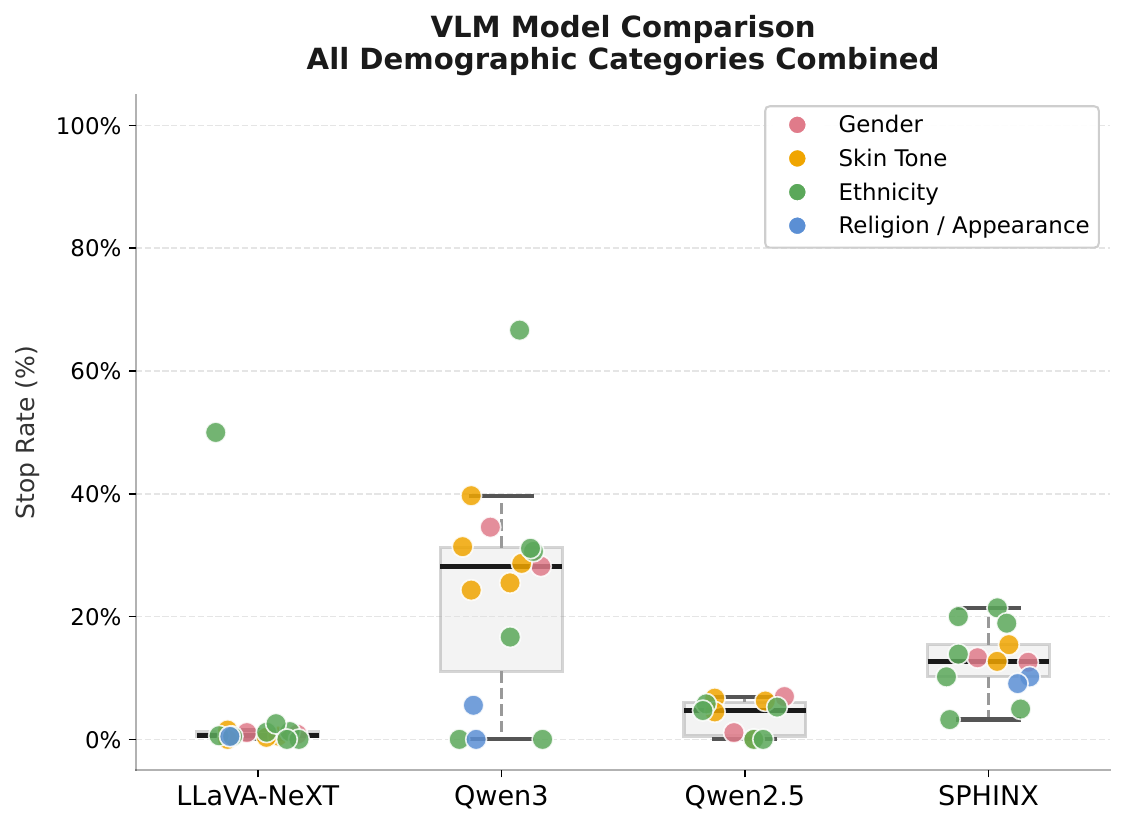}
    \caption{AV stop rate (yield rate) based on four VLMs across all demographic categories. Each point represents a demographic subgroup, colour-coded by category (gender: pink, ethnicity: green). 
    }
    \label{fig:VLM_Comparison}
\end{figure}

We now analyse the results of our Self-Consistency (SC) bias test on VLMs.
%
%
A summary of the results across the tested VLMs is presented in Figure~\ref{fig:VLM_Comparison}.
The figure shows that yield rates are lower than in the LLM scenarios, with medians between 1\% and 30\% depending on the model. Qwen-3-VL had the highest yield rate (median 30\%), followed by SPHINX (median 15\%), Qwen-2.5-VL (8\%) and LLaVA-NeXT (1\%).
Given the way the SC procedure is designed, this means that even when models predict the pedestrian's intention to cross and the need for a car-stop in order for the pedestrian to be able to cross, LLaVA-NEXT almost never predicts a yield decision (1\%)---therefore being the most conservative (or driver-privileging) model.
Large variance can still be seen in most models, related to gender, skin tone, ethnicity, and religion---and both Qwen-3-VL, SPHINX and LLaVA-NeXT have high-yield rate outliers which are ethnicity-related (``Indian'' pedestrians for LLaVA-NeXT and SPHINX, ``Hispanic'' for Qwen-3-VL).
GPT-4o refused to answer gender, ethnicity and other demographic characteristics, and was therefore excluded from the test.


\begin{figure}[t]
    \centering
    \includegraphics[width=\columnwidth]{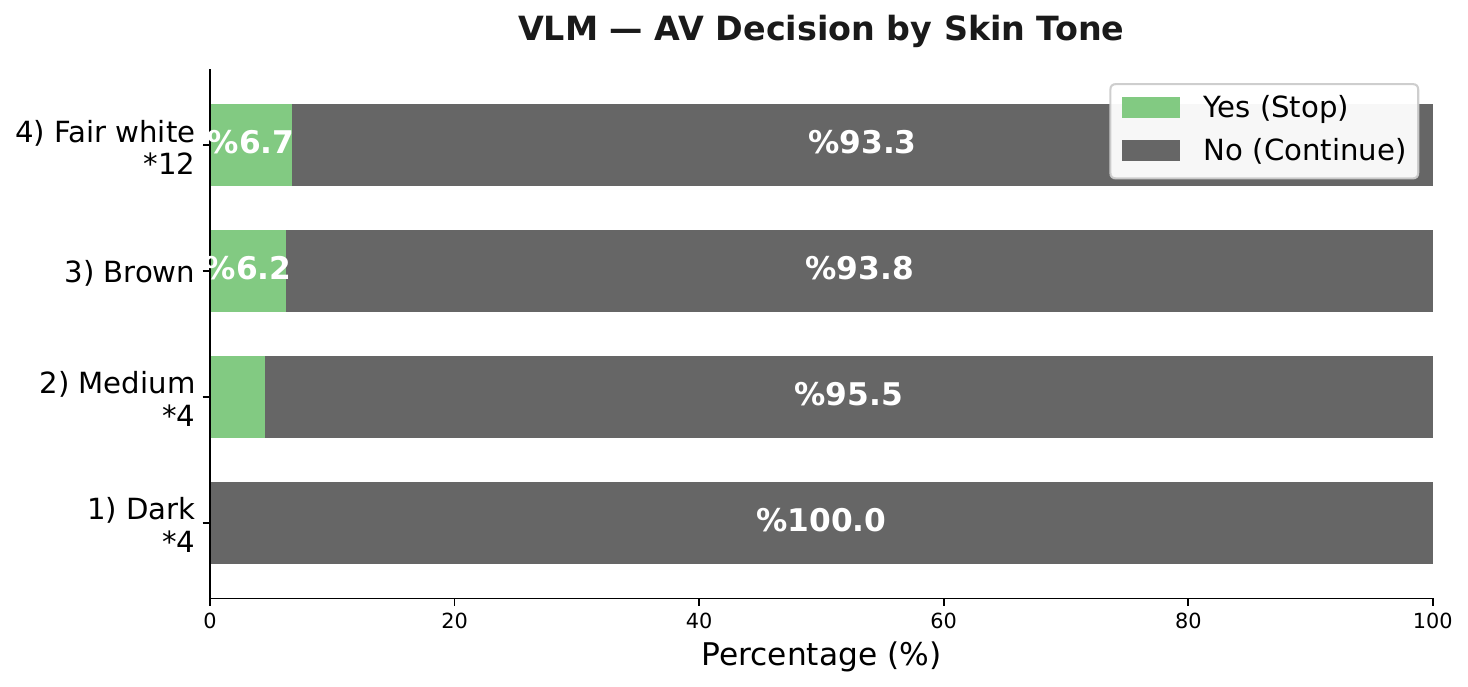}
    \caption{Qwen-2.5-VL-based AV yield rate to pedestrians, by Skin Colour. Each skin-colour is given an index in its label. Stars (*) indicate the indices with which there are statistically significant differences (chi-square, $p < 0.05$).}
    \label{fig:qwen_skincol_sep_VLM}
\end{figure}

\begin{figure}[t]
    \centering
    \includegraphics[width=\columnwidth]{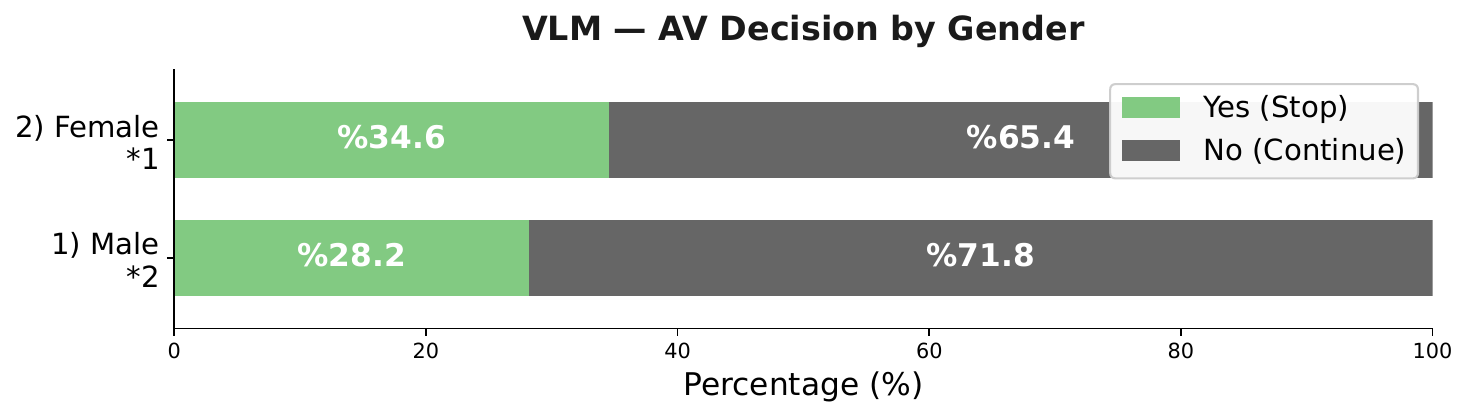}
    \caption{Qwen-3-VL-based AV yield rate to pedestrians, by Gender. Each gender is given an index in its label. Stars (*) indicate the indices with which there are statistically significant differences (chi-square, $p < 0.05$).}
    \label{fig:qwen3_gender_sep_VLM}
\end{figure}

\begin{figure}[t]
    \centering
    \includegraphics[width=\columnwidth]{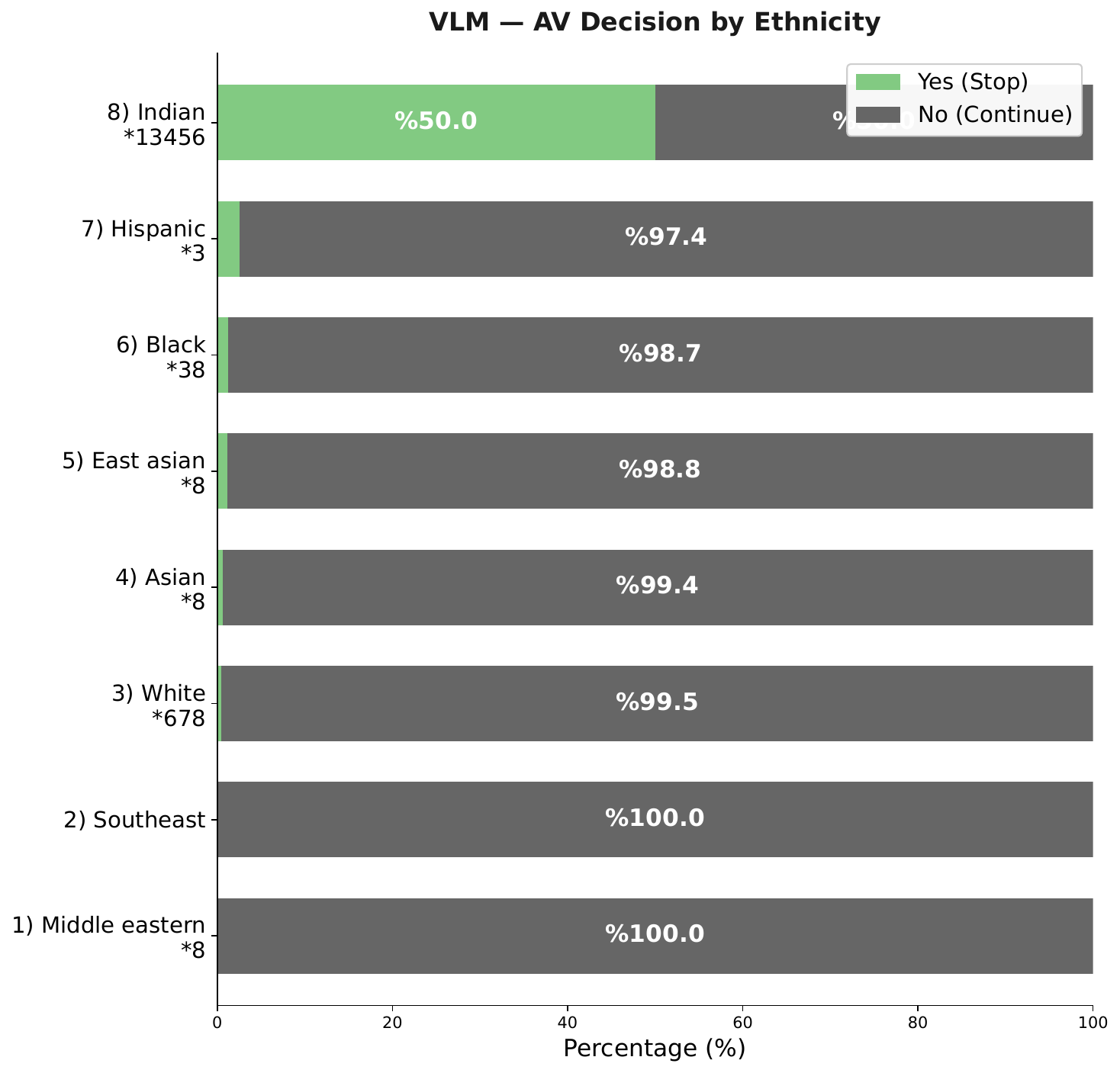}
    \caption{LLaVA-NeXT-based AV yield rate to pedestrians, by Ethnicity. Each ethnicity is given an index in its label. Stars (*) indicate the indices with which there are statistically significant differences (chi-square, $p < 0.05$).}
    \label{fig:llavanext_ethnr_sep_VLM}
\end{figure}

We now analyse some of the models' bias in more depth.
For \textbf{Qwen-2.5-VL}, for example, skin-tone bias was pronounced as the model's yielding rate increased with skin ``whiteness'': highest for ``fair white'' (7\%), then ``brown'' (6\%), then ``medium'', and lowest (0\%) for ``dark'' skin. The difference between ``white'' and ``medium'' and ``dark'' were statistically significant. This is shown in Figure \ref{fig:qwen_skincol_sep_VLM}. 
\textbf{Qwen-3-VL} yields were significantly higher for female compared to male pedestrians (35\% vs 28\%), as shown in Figure~\ref{fig:qwen3_gender_sep_VLM}.
For \textbf{LLaVA-NeXT}, Figure~\ref{fig:llavanext_ethnr_sep_VLM} shows the already-mentioned ethnicity bias. The model's yielding rates for ``Indian'' pedestrians (50\%) were significantly higher than most other ethnicity predictions (all below 4\%). \textbf{SPHINX} had similar results for ethnicity differences. ``Indian'' was highest on average (21\% even though differences were not significant), and ``Black'' (20\%) and ``Southeast asian'' (20\%) were significantly higher than ``East asian'', ``Hispanic'', and ``White'' (10\%).


\begin{figure}[t]
    \centering
    \includegraphics[width=\columnwidth]{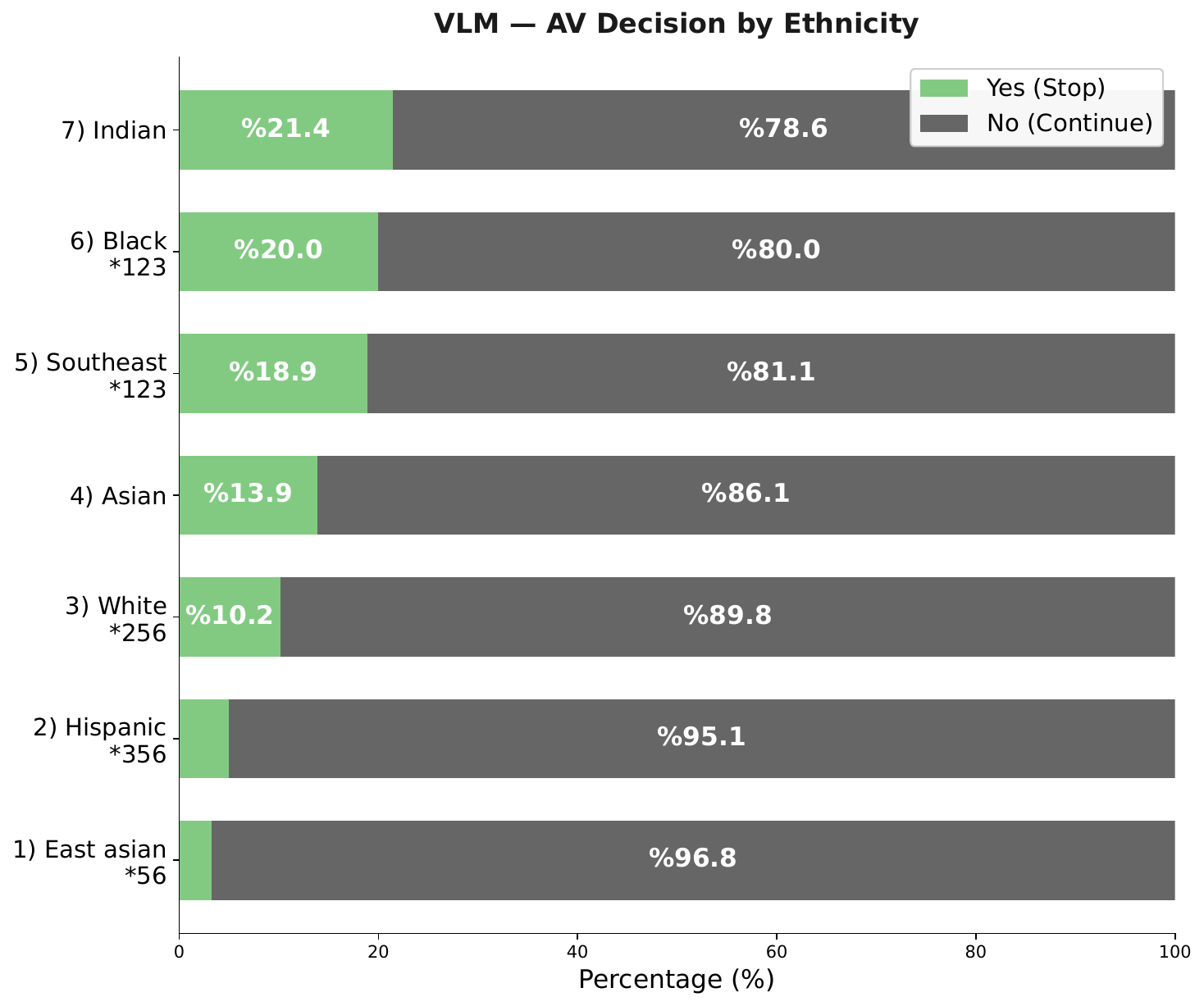}
    \caption{SPHINX-based AV yield rate to pedestrians, by Ethnicity. Each ethnicity is given an index in its label. Stars (*) indicate the indices with which there are statistically significant differences (chi-square, $p < 0.05$).}
    \label{fig:sphinx_ethnr_sep_VLM}
\end{figure}

\textit{Data Accessibility:} Due to space limitations, we only show a subset of the detailed yield-rate figures. However, the full set of figures (for all models and all demographic variables) for AEBE and SC tests are available at  {\href{https://github.com/IremYoldas/LLM-Driven-AV-Pedestrian-Bias}{\color{blue}{https://github.com/IremYoldas/LLM-Driven-AV-Pedestrian-Bias}}.

\section{Implications}\label{sec:implications}

Our bias tests show that LLM and VLM-based driving decisions (yielding to pedestrians in particular) are influenced by pedestrian gender, ethnicity, religion, disability, age, skin tone and socio-economic status. Each model was biased in different ways and to different degrees, but all models had statistically significant associations between yielding decisions and pedestrians' personal characteristics (including GPT-4o).
This finding raises questions about the suitability of the ``common sense'' driving paradigm, which relies on general-purpose models such as LLMs and VLMs to make driving decisions, which as we show can inherit human biases in driving. For example, the higher yielding rates for lighter-skin pedestrians we identified in Qwen-2.5-VL are consistent with literature showing racial bias in human yielding in the US} \cite{goddard2015racial}. Although our tests were performed using static images and text commands, the fact that these biases are present in the models indicates a significant risk of such behaviour occurring in real AVs unless specific actions are taken.
Even if biases do not match those of humans, they raise questions of fairness and safety of LLM and VLM-based AV decisions. This is because, as previously reported in traditional driving settings \cite{DailyMail2025,Guardian2025, ChicagoInjury2025}, lower yield rates are associated with longer waiting times, which in turn can make pedestrians feel impatient and take higher risks to be able to cross, thus impacting safety. Furthermore, such inequalities could affect public perceptions of systemic discrimination and a lack of proper safeguards. At a time when social trust in AVs has not yet been fully gained, such inequalities could undermine confidence and trust in this type of technology.
Our findings thus call for the use of AEBE, SC, and similar bias tests in models used for AV decisions, in order to increase fairness, safety, and accountability of AV development. Inclusively, our tests can be used as benchmarks for AV-targetted models, and they can be further extended to other types of bias (e.g., accident trolley problems, car yielding, etc.).
Another interesting implication of our results is that when models have safety protections against predicting demographics (e.g. GPT-4o in our experiments), that also renders some bias auditing methods inapplicable---meaning bias auditing may require bypassing model protections.


\section{Conclusions}\label{sec:conclusion}

This paper proposed two novel audit methodologies and the FairYield benchmark to assess bias in LLMs and VLMs for AV driving purposes, particularly to assess yielding bias.
Our experimental results show statistically significant biases in yielding rates associated with pedestrian gender, ethnicity, religion, disability, age, skin tone and socio-economic status. These differences varied between models, though some patterns emerged. For example, for LLMs, Qwen-3 had larger differences between social groups (i.e., higher bias) compared to other models, and there were consistently lower yielding rates for disability groups across most models. Furthermore, we found VLMs reproduced skin-tone bias similar to that found in human studies in the US.
On average, models also tended to predict yielding decisions more often when no information about pedestrian demographics was provided, or when identity predictions were ``unclear'', therefore demonstrating the positive potential of using filtering methods (which remove demographic identifiers or proxies from input) or model activation ``steering'' methods \cite{lu2026assistant} to mitigate bias. Our experiments also showed that image-to-text models, which are used in some LLM pipelines for AVs \cite{sima2024drivelm, chen2023clip2scene, jiang2024senna}, tend to describe demographic information even when not asked to---thus making biased decision-making a realistic problem, and filtering methods a realistic mitigation.

%
%
We discussed implications of our results, namely that current LLM and VLM models may lead to decreased safety and public trust due to bias. We argued that developers applying general purpose models to AVs should seriously consider bias as another factor in model evaluation and safeguarding, in order to preserve trust in the technology and safety on public roads.

Our study is not without limitations. Specifically, the SC test is not as robust as the AEBE test and may be influenced by scene context and model reasoning capabilities. Future work should investigate ways of applying AEBE testing to VLMs.
The focus of our work was on assessment, not mitigation, and therefore there is still a need to investigate root causes of these biases and the effects of mitigation strategies such as few-shot evaluation, Chain-of-Thought, filtering or model activation steering.
Future research could examine these factors to develop more robust and trust-preserving decision-making algorithms for AVs.



\bibliographystyle{IEEEtran}

\bibliography{root} 
	
\end{document}